\documentclass[runningheads]{llncs}
\usepackage[T1]{fontenc}

\usepackage{graphicx}
\usepackage{booktabs}
\usepackage{multirow}
\usepackage{amsmath}
\usepackage{hyperref}
\usepackage[nolist,nohyperlinks]{acronym}
\usepackage{comment}
\usepackage{orcidlink}
\usepackage{wrapfig}

\usepackage{xcolor}
\usepackage{lipsum}
\usepackage{color}

\begin{document}
\begin{acronym}
    \acro{LLM}{Large Language Model}
    \acro{MCP}{Model Context Protocol}
    \acro{QA}{Question Answering}
    \acro{NL}{Natural Language}

\end{acronym}

\title{From Single to Multi-Agent Reasoning: Advancing GeneGPT for Genomics QA}
\titlerunning{GenomAgent: Multi-Agent Reasoning for Genomics QA}

\author{Kimia Abedini\inst{1}\orcidlink{0009-0007-2586-4920} \and 
Farzad Shami\inst{2}\orcidlink{0009-0004-8174-0082} \and
Gianmaria Silvello\inst{1}\orcidlink{0000-0003-4970-4554}}

\institute{University of Padua, Italy \and Aalto University, Finland}

\maketitle 

\begin{abstract}
Comprehending genomic information is essential for biomedical research, yet extracting data from complex distributed databases remains challenging. Large language models (LLMs) offer potential for genomic \acf{QA} but face limitations due to restricted access to domain-specific databases. GeneGPT is the current state-of-the-art system that enhances LLMs by utilizing specialized API calls, though it is constrained by rigid API dependencies and limited adaptability. We replicate GeneGPT and propose \texttt{GenomAgent}, a multi-agent framework that efficiently coordinates specialized agents for complex genomics queries. Evaluated on nine tasks from the GeneTuring benchmark, \texttt{GenomAgent} outperforms GeneGPT by 12\% on average, and its flexible architecture extends beyond genomics to various scientific domains needing expert knowledge extraction.


\vspace{0.6em}
\hspace{3.5em}\includegraphics[width=1.25em,height=1.25em]{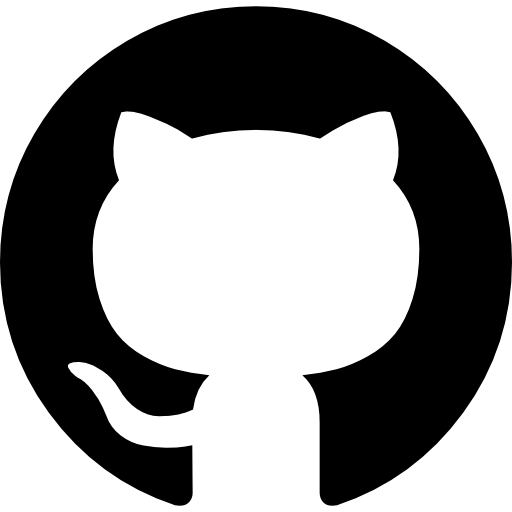}\hspace{.3em}
\parbox[c]{\columnwidth}
{
    \vspace{-.55em}
    \href{https://kimia-abedini.github.io/Genom-Agent/}{\nolinkurl{https://kimia-abedini.github.io/Genom-Agent/}}
}
\vspace{-1.2em}
\end{abstract}

\section{Introduction}
\label{sec:introduction}
{\acp{LLM} have shown remarkable potential in \ac{QA} tasks and have recently gained traction in genomic \ac{QA} applications~\cite{li2024developing,ali2025large}.
A notable and widely cited example is GeneGPT~\cite{jin2024genegpt}, which currently represents the state-of-the-art for genomic \ac{QA} tasks by successfully augmenting \acp{LLM} with external domain-specific APIs through in-context learning~\cite{brown2020language} and tool integration \cite{shen2024llm}. 
GeneGPT operates as a single-agent architecture\cite{masterman2024landscape} where an \ac{LLM} is guided through carefully constructed prompts containing API documentation and examples, with inference managed sequentially through a single forward loop of API calls and result processing. Despite its effectiveness in achieving high accuracy on genomic benchmarks, GeneGPT's architecture exhibits several limiting characteristics that constrain its scalability and adaptability. The system's rigid dependency on specific API formats makes it fragile when interfacing with evolving tools, while its reliance on extensive context windows can lead to attention dilution and reduced focus on the original query~\cite{liu2023lost,hsieh2024found}. Furthermore, the sequential processing approach struggles with multi-turn conversations~\cite{laban2025llms} where context drift becomes problematic, and the stop-token mechanisms for API call extraction lack the robustness needed for integration with newer \acp{LLM}. 

In response to these limitations and building upon recent advances in multi-agent \ac{LLM} systems \cite{cemri2025multi}, we propose a novel multi-agent architecture that addresses these efficiency bottlenecks through specialized agent coordination and dynamic task decomposition.
We first conduct a GeneGPT reproducibility study and adapt the system to more recent \acp{LLM} to identify key limitations. Second, we introduce \texttt{GenomAgent}, a multi-agent framework that extends GeneGPT’s capabilities.
Experimental results show \texttt{GenomAgent} achieves an average performance score of 0.93 (+12\%  over GeneGPT's 0.83) while reducing computational costs by 79\% (\$2.11 vs. \$10.06 total) across the GeneTuring benchmark~\cite{hou2023geneturing}. 

The remainder of the paper is organized as follows: Section~\ref{sec:introduction} reviews GeneGPT, Section~\ref{sec:reproduction} details GeneGPT replication, Section~\ref{sec:reproduction} describes \texttt{GenomAgent}, Section~\ref{sec:experiments} presents the experiments, and Section~\ref{sec:conclusion} provides some final remarks.

\section{GeneGPT}\label{sec:genegpt}
GeneGPT\cite{jin2024genegpt} is a domain-specific system that enhances LLMs by integrating a tool-augmented architecture to connect \ac{NL} queries with structured genomics databases. It utilizes in-context learning~, enabling the LLM to dynamically generate and execute API calls to external resources, thus allowing real-time data retrieval and synthesis. This approach overcomes the limitations of static knowledge repositories in pre-trained models and demonstrates the extended utility of LLMs in specialized fields by ensuring access to up-to-date, structured data, while retaining their NLP capabilities for scientific \ac{QA}.

GeneGPT employs a specialized prompting strategy that leverages the code completion capabilities of \acp{LLM}. It is based on OpenAI Codex\cite{chen2021evaluating}, and the prompt structure includes task instructions, relevant API documentation for E-utils and BLAST~\cite{schuler199610,altschul1990basic,ncbi2015database}, in-context learning examples, and the target question. GeneGPT uses the special symbol ``$\rightarrow$'' as a stop token to identify API calls. When the LLM generates text containing this symbol, the system: (1) extracts the URL using a regex pattern; (2) executes the API call; and (3) appends the API result to the prompt. The model then continues generation, repeating steps 1-3 for any additional API calls, until the termination token  ``$\backslash \texttt{n}\backslash \texttt{n}$'' is detected. 
Then, the \ac{LLM} generates the final answer using the retrieved results and in-context understanding of the examples.

GeneGPT was developed in four configurations: full, slim, turbo, and lang. In full settings, the system incorporates complete API documentation and four examples, while slim uses only two examples. The turbo configuration replaces Codex~ with \texttt{GPT-3.5-turbo-16k}, and lang implements the ReAct framework~\cite{yao2022react}. The system was evaluated on nine tasks in the GeneTuring benchmark. Based on the experimental results, GeneGPT achieves state-of-the-art performance with an average performance score of 0.83, which substantially outperforms baselines as Bing Chat (0.44), BioMedLM~\cite{luo2022biogpt} (0.08), and GPT-3 (0.16).

GeneGPT performance was assessed across multiple evaluation metrics designed for different tasks within the GeneTuring benchmark. These include exact match accuracy for nomenclature tasks, recall for association tasks, and task-specific scoring for alignment tasks. While individual task metrics employ different evaluation criteria and cannot be directly compared inter-task due to varying task complexity and requirements, all metrics are normalized in $[0,1]$, enabling uniform interpretation. For comparative analysis, following an established approach in multi-task evaluation~\cite{wang2019,wang2018}, we report a macro-averaged performance score computed as the arithmetic mean across all task-specific metrics, providing a singular measure of overall system accuracy while acknowledging that this aggregate metric represents a simplified view of the system's diverse capabilities across heterogeneous genomics \ac{QA} tasks.

\section{Reproducibility of GeneGPT} \label{sec:reproduction}
To understand GeneGPT's operational principles and identify improvement opportunities, we conducted a reproducibility study. The original system relied on \texttt{code-davinci-002} and \texttt{GPT-3.5-turbo-16k}, which were deprecated in 2023 and 2024, respectively\footnote{\url{https://platform.openai.com/docs/deprecations}}. 
We selected \texttt{GPT-4o-mini} as the replacement model due to its performance, cost efficiency, and \textcolor{black}{current stability}.
We implemented two compatible configurations: \texttt{turbo} and \texttt{lang}. The original paper for the \texttt{lang} setting mentions only \emph{LangChain} as the orchestration framework without detailing its implementation. Due to substantial changes and deprecation in favor of LangGraph\footnote{\url{https://langchain-ai.github.io}}, we opted for LangGraph for this configuration.
We preserve GeneGPT's core design based on the stop-token interaction mechanism.

\begin{table}[t]
    \Large
    \centering
    \caption{Results of the reproducibility of GeneGPT on the  GeneTuring Benchmark.
    }\label{tab1}
    \resizebox{0.99\linewidth}{!}{\begin{tabular}{|l|cc|ccc|cc|cc|}
        \hline
        \multirow{3}{*}{Model} & 
        \multicolumn{2}{c|}{Nomenclature} & 
        \multicolumn{3}{c|}{GenomicLocation} & 
        \multicolumn{2}{c|}{FunctionalAnalysis} & 
        \multicolumn{2}{c|}{SequenceAlignment}  \\
        \cline{2-10}
        \rule{0pt}{5ex}
        & 
        \shortstack[b]{Gene\\Alias} & \shortstack[b]{Name\\Conv.}  & 
        \shortstack[b]{SNP\\Assoc.} & \shortstack[b]{Gene\\Loc.} & \shortstack[b]{SNP\\Loc.} & 
        \shortstack[b]{Disease\\Assoc.} & \shortstack[b]{Protein\\Genes} & 
        \shortstack[b]{DNA to\\Human} & \shortstack[b]{DNA to\\Species} \\
        \hline
        GeneGPT Turbo & 0.64 & 1.00 & 0.96 & 0.54 & 0.98 & 0.63 & 0.96 & 0.42 & 0.88 \\
        Reproduced & 0.68 & 0.98 & 0.90 & 0.54 & 0.92 & 0.56 & 0.80 & 0.07 & 0.62 \\
        \hline
        \textbf{Relative diff} & 6.25\% & -2.00\% & -6.25\% & 0.00\% & -6.12\% & -11.11\% & -16.67\% & -83.33\% & -29.55\%\\
        \hline\hline 
        GeneGPT Lang & 0.76 & 0.02 & 0.90 & 0.54 & 0.74 & 0.39 & 0.90 & 0.06 & 0.54 \\
        Reproduced & 0.76 & 0.92 & 1.00 & 0.72 & 1.00 & 0.76 & 1.00 & 0.31 & 0.54 \\
        \hline
        \textbf{Relative diff} & 0.00\% & 4500\% & 11.11\% & 33.33\% & 35.14\% & 94.87\% & 11.11\% & 416.67\% & 0.00\%\\
        \hline
    \end{tabular}}
\end{table}

During the reproduction process, we encountered two main challenges. First, GPT-4o-mini did not consistently follow the URL generation format required by GeneGPT's extraction pipeline. We addressed this by explicitly prompting the model to use the desired format. 
Second, the original implementation used context truncation to avoid exceeding length limits, which hindered HTML data extraction by discarding critical information. We removed this limit with GPT-4o-mini's larger context window.
Unlike the original system's single-token outputs, the reproduced system often requires manual extraction from multi-sentence responses before automatic evaluation.

For the reproducibility analysis, we employ the GeneTuring Benchmark, which encompasses 12 distinct tasks, each comprising 50 question-answer pairs. We approached 9 of these GeneTuring tasks, replicating the original GeneGPT paper. These selected tasks are grouped into four main subcategories: (1) nomenclature inquiries, focusing on gene aliases and name transformations; (2) genomic location inquiries, examining the positioning of genes and SNPs and their interrelations; (3) functional analysis inquiries, investigating aspects such as gene-disease associations and the genes responsible for protein coding; and (4) sequence alignment inquiries, which involve mapping DNA sequences to the human genome and comparing them across various species.

Table~\ref{tab1} presents the reproduced results. Our reproduced system consistently shows improvements in the \texttt{lang} setting; these gains show that correct implementation of ReAct architecture with newer models can increase performance. However, in \texttt{turbo} settings, we observed high variation and degradation as a result of the non-compatibility of stop-token processing with general-purpose \acp{LLM}.
We manually reviewed and categorized all the mistakes made by the system we replicated into three distinct types: E1: incomplete data coverage, where correct answers do not exist in NCBI; E2: stop-token parsing failures, where \ac{LLM} does not generate API calls in the expected format; E3: context loss, where large API responses cause \ac{LLM} to lose focus on the original question. Our results suggest that the reproduced \texttt{turbo} setting causes errors due to E2, where the system gets stuck in a loop, and ultimately, no results are achieved. In contrast, in \texttt{lang} mode, the most dominant errors are related to E1 and E3.

\section{GenomAgent}
\label{sec:genomagent}

\begin{figure}[t!]
  \centering
  \includegraphics[width=\textwidth]{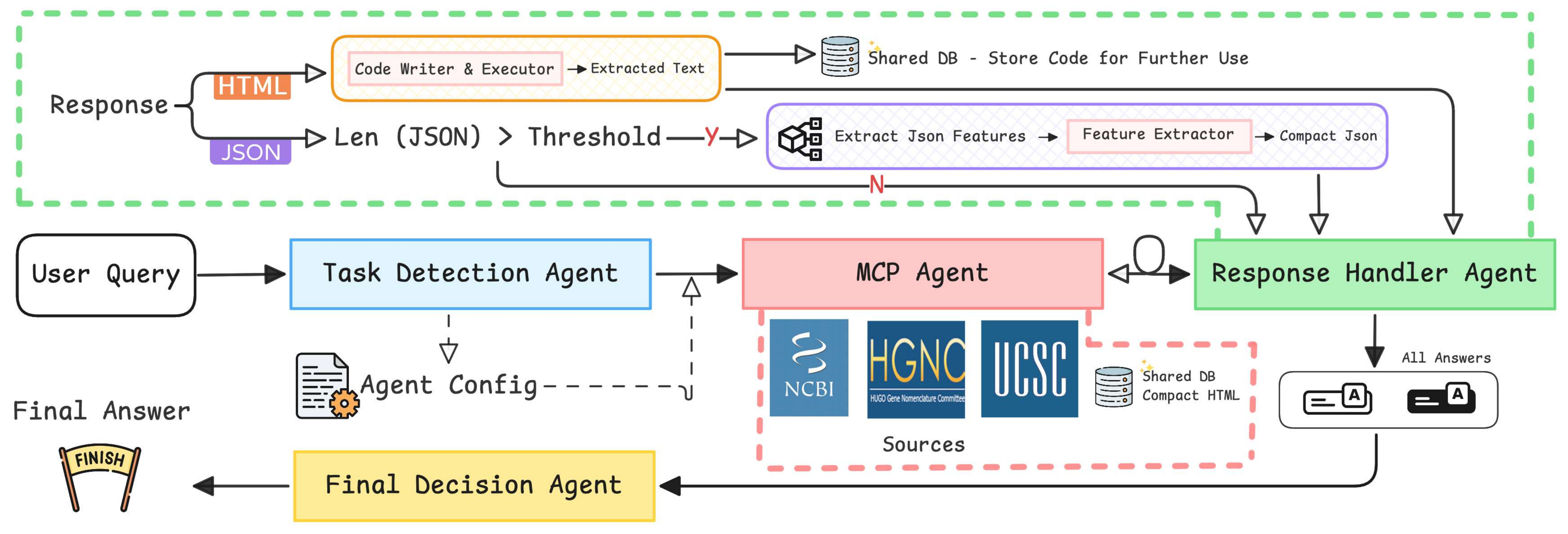}
  \caption{\texttt{GenomAgent} multi-agent architecture and workflow.}
  \label{fig:genomagent}
\end{figure}

We present \texttt{GenomAgent}, a multi-agent architecture (see Figure \ref{fig:genomagent}) that extends beyond single-agent approaches for biomedical \ac{QA}. The system uses multiple specialized agents to handle questions through a coordinated workflow and enable flexible interaction with various biomedical APIs and DBs.

\texttt{GenomAgent} implements a hierarchical multi-agent architecture comprising four core processing agents and three specialized utility agents. The \textbf{Task Detection Agent} serves as the initial query router, performing intent classification to determine appropriate processing workflows based on predefined configuration schemas. The \textbf{Multi-source Coordination Protocol (MCP) Agent} orchestrates parallel API interactions across heterogeneous biomedical databases (NCBI~\cite{ncbi2015database}, HGNC~\cite{seal2023genenames}, UCSC~\cite{perez2025ucsc}), implementing asynchronous query dispatch and response aggregation protocols.
The \textbf{Response Handler Agent} processes heterogeneous API responses through dual processing pipelines: (1) JSON responses undergo threshold-based evaluation, triggering the \textbf{Feature Extractor Agent} for schema summarization when size limits are exceeded, and (2) HTML responses activate the \textbf{Code Writer Agent} to generate targeted extraction scripts executed by the \textbf{Code Executor Agent}. Generated extraction code is cached in a shared repository to enable reuse and reduce computational overhead. The \textbf{Final Decision Agent} performs multi-source response synthesis using consensus-based aggregation algorithms to generate coherent answers.

Built on the Google Agent Development Kit, \texttt{GenomAgent} addresses three critical limitations identified in GeneGPT: (1) \textit{source diversity} through multi-database querying to reduce information gaps, (2) \textit{modular processing} via specialized agents to handle heterogeneous response formats, and (3) \textit{adaptive extraction} through dynamic code generation for complex data structures. This architecture enables parallel processing, reduces context window constraints, and provides fault tolerance through distributed task execution.

\section{Experiments} \label{sec:experiments}

\texttt{GenomAgent} evaluation follows the same experimental setup as our reproducibility study with enhanced precision improvements. Task specific evaluation metrics include: exact matching for nomenclature and genomic location tasks; recall calculation based on exact gene matches for gene-disease associations; vocabulary-mapped exact matching for cross-species DNA alignment (mapping Latin to common names, e.g., ``Homo sapiens'' to ``human''); and partial scoring for human genome alignment, awarding 0.5 points for correct chromosome identification with incorrect positions (e.g., \texttt{chr8:708–882} vs. \texttt{chr8:120–121}).

Our experimental protocol differs from GeneGPT in two key aspects: (1)
expanded vocabulary mappings to accommodate updated NCBI species annotations, and (2) enhanced partial scoring that calculates sequence-level similarity for both start and end positions in alignment tasks. 
\textcolor{black}{We applied identical evaluation protocols to both GeneGPT and \texttt{GenomAgent} to ensure fair comparison. The expanded vocabulary mappings and partial scoring mechanisms were applied to both systems when evaluating on the GeneTuring benchmark.}

\begin{table}[th]
    \Large
    \centering
    \caption{Performance and cost (\$) on GeneTuring. Best existing models are underlined; bottom row shows \texttt{GenomAgent}'s improvement over best baseline.}\label{tab2}
    \resizebox{0.99\linewidth}{!}{\begin{tabular}{|l|cc|cc|cc|cc|cc|}
        \hline
        \shortstack[b]{Model\\~} & 
        \multicolumn{2}{c|}{\shortstack[b]{Nomenclature\\~}} & 
        \multicolumn{2}{c|}{\shortstack{Genomic\\Location}} & 
        \multicolumn{2}{c|}{\shortstack{Functional\\Analysis}} & 
        \multicolumn{2}{c|}{\shortstack{Sequence\\Alignment}} & 
        \multicolumn{2}{c|}{\shortstack[b]{Overall\\~}}  \\
        \cline{2-11}
         & Score$\uparrow$ & Cost$\downarrow$ & Score$\uparrow$ & Cost$\downarrow$ & Score$\uparrow$ & Cost$\downarrow$ & Score$\uparrow$ & Cost$\downarrow$ & Avg$\uparrow$ & Total(\textdollar)$\downarrow$ \\
        \hline
        GeneGPT Full & 0.90 & 2.19 & 0.87 & \underline{2.36} & 0.76 & 2.13 & 0.65 & 1.69 & 0.80 & 11.14 \\
        GeneGPT Slim & \underline{0.92} & \underline{1.63} & \underline{0.88} & 2.52 & \underline{0.84} & \underline{1.67} & \underline{0.66} & \underline{1.74} & \underline{0.83} & \underline{10.06}\\
        GeneGPT Turbo & 0.82 & 4.39 & 0.82 & 4.73 & 0.80 & 4.25 & 0.65 & 3.39 & 0.78 & 16.76 \\
        GeneGPT Lang & 0.39 & n/a & 0.73 & n/a & 0.65 & n/a & 0.30 & n/a & 0.54 & n/a\\
        \hline\hline 
        \texttt{GenomAgent} (Ours) & \textbf{0.98} & \textbf{0.43} & \textbf{0.98} & \textbf{0.88} & \textbf{0.89} & \textbf{0.25} & \textbf{0.85} & \textbf{0.55} & \textbf{0.93} & \textbf{2.11}\\
        \hline
        \textbf{Improvement (\%)} & 6.5\% & 73.6\% & 11.4\% & 62.7\% & 4.8\% & 85.0\% & 28.8\% & 68.4\% & 12.0\% & 79.0\%\\
        \hline
    \end{tabular}}
    \label{tbl:performance}
\end{table}
Furthermore, we quantify the computational cost for each task. This is achieved by tracking the number of input and output tokens and applying real-model pricing to derive the total cost per task.


\begin{figure}[th]
    \centering
    \includegraphics[width=0.82\textwidth]{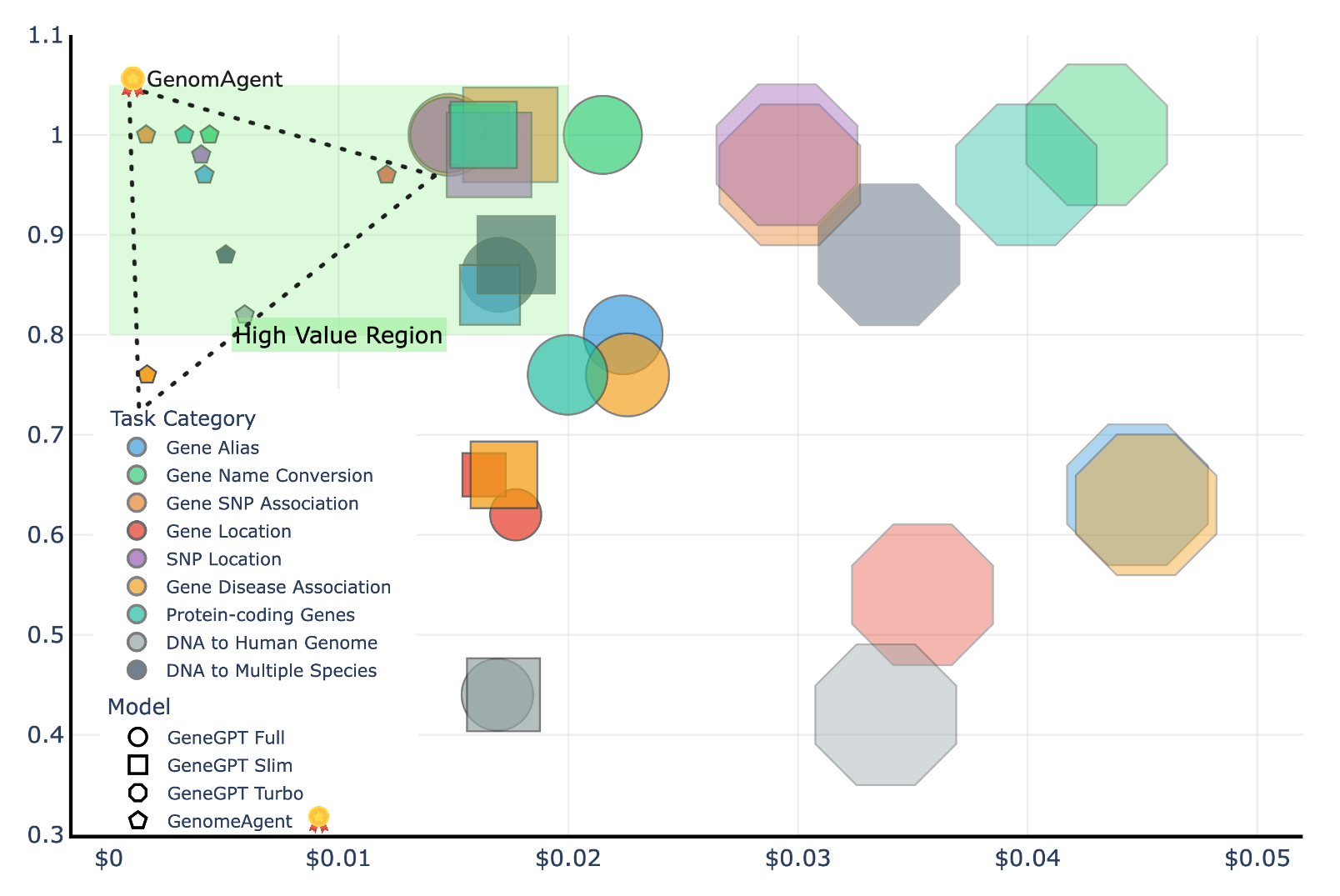}
    \caption{Performance-cost tradeoff on GeneTuring. Bubble size shows normalized cost; High Value Region shows optimal performance at minimal cost.}
  \label{fig:performance}
\end{figure}

Table~\ref{tbl:performance} reports the performance and cost of \texttt{GenomAgent} in GeneTuring tasks compared to GeneGPT's main results. \texttt{GenomAgent} achieves substantial improvements in both performance and computational efficiency. 
Our model attains an average score of 0.93, exceeding the best-performing GeneGPT model (0.83).
In simple tasks (nomenclature and genomic location), our system achieves near-perfect performance with a score of 0.98, surpassing GeneGPT-slim's scores of 0.92 for nomenclature and 0.88 for genomic location.
Most notably, in alignment tasks, which are the most challenging task for GeneGPT, we achieve a remarkable 28.8\% improvement.
Computational cost analysis reveals even more striking improvements. \texttt{GenomAgent} costs only \$2.11 total in all tasks (79.0\% reduction from best-performing GeneGPT (\$10.06)). 

In addition, as shown in Figure~\ref{fig:performance}, \texttt{GenomAgent} is the optimal selection, as it achieves a high score at minimal computational expense. 

\section{Final Remarks and Future Work}\label{sec:conclusion}
In this study, we reproduce GeneGPT~\cite{jin2024genegpt} to pinpoint three critical bottlenecks: (i) limited data coverage, (ii) parsing failures, and (iii) context loss in multi-turn queries. We then introduce \texttt{GenomAgent}, a hierarchical multi-agent framework that orchestrates parallel API queries, dynamic data extraction, and consensus-based response synthesis. Evaluated on the GeneTuring benchmark, \texttt{GenomAgent} achieves a 12\% increase in average performance (0.93 vs.\ 0.83) and a 79\% reduction in computational cost (\$2.11 vs.\ \$10.06). Sequence alignment tasks see the largest gains (28.8\%), driven by multi-source retrieval and adaptive partial scoring. Unlike GeneGPT’s rigid single-agent design, \texttt{GenomAgent}’s modular agents seamlessly adapt to new LLMs and evolving database schemas. These results demonstrate that coordinated multi-agent orchestration can deliver both superior accuracy and substantial resource efficiency for genomic question answering.

Looking ahead, our results suggest several promising research directions: \textcolor{black}{First, we acknowledge that the 12\% average improvement cannot be cleanly attributed to specific architectural choices without systematic ablation analysis. Decomposing components through controlled experiments that isolate individual elements can demonstrate the contribution of each architectural component. Second, our evaluation is limited to the GeneTuring benchmark. This restricted scope prevents us from fully validating \texttt{GenomAgent}'s generalizability across diverse genomic QA tasks. Third, investigating hybrid approaches that combine the efficiency of single-agent systems for simple queries with multi-agent coordination for complex tasks could optimize the performance-cost tradeoff. Fourth, the development of automated prompt optimization techniques for agent-specific instructions could further reduce the manual effort required for system configuration. Finally, extending our comparative analysis to include emerging state-of-the-art frameworks such as \cite{chen2025beyond} will enable benchmarking of our system's capabilities against the latest advances. We will investigate all these dimensions in the planned future work.}

\section*{Acknowledgments} 
This work is partially supported by the HEREDITARY Project, as part of the European Union's Horizon Europe research and innovation programme under grant agreement No GA 101137074.

\section*{Disclosure of Interests}
The authors have no competing interests to declare that are relevant to the content of this article.

%
%
%
\newpage
\bibliographystyle{splncs04}
\bibliography{references}
%




\end{document}